\documentclass[runningheads]{llncs}
\usepackage{graphicx}
\usepackage{comment}
\usepackage{amsmath,amssymb} 
\usepackage{color}
\usepackage{appendix}
\usepackage{nicefrac}
\usepackage{epsfig}
\usepackage{graphicx}
\usepackage[table]{xcolor}
\usepackage{amsmath}
\usepackage{amssymb}
\usepackage[font=small]{caption}

\usepackage{pifont}

\usepackage{cuted}
\usepackage{capt-of}

\usepackage{algorithm}
\usepackage{algorithmicx}
\usepackage[noend]{algpseudocode}
\usepackage{booktabs}
\usepackage{multirow}
\usepackage{subfig}
\usepackage{diagbox} 
\usepackage{arydshln} 
\usepackage{tabularx}
\usepackage{xspace}
\usepackage[numbers,sort]{natbib} 

\DeclareCaptionLabelFormat{showtable}{\tablename~\thetable(#2):}

\definecolor{lightyellow}{RGB}{255,255,170}

\definecolor{lightblue}{RGB}{0,100,255}

\definecolor{boxred}{RGB}{225, 104, 82}
\definecolor{boxblue}{RGB}{103, 136, 237}

\usepackage[pagebackref=true,breaklinks=true,letterpaper=true,colorlinks,bookmarks=false,citecolor=lightblue]{hyperref}

\def\newpara{\vspace{3pt}}

\usepackage{amsmath}
\DeclareMathOperator*{\argmax}{arg\,max}

\newcommand{\sfig}[1]{Figure~\ref{#1}}

\newcommand{\xpar}[1]{\vspace{-2.5mm}\paragraph{\normalfont\bf #1}\ \ }

\newcommand{\modelname}[0]{{\mbox{LWTNet}}\xspace} 
\newcommand{\figvspace}[0]{\vspace{-6mm}}
\newcommand{\mysect}[1]{\vspace{-2.75mm}\section{#1}\vspace{-1.5mm}}
\newcommand{\mysubsect}[1]{\vspace{-2.5mm}\subsection{#1}\vspace{-1.5mm}}

\def\eg{\emph{e.g.}}

\def\etal{\emph{et al.}}

\newcommand{\ao}[1]{}
\newcommand{\daffy}[1]{}
\newcommand{\az}[1]{}
\newcommand{\joon}[1]{}

\newcommand{\webpage}{\href{http://www.robots.ox.ac.uk/~vgg/research/avobjects}{webpage}\xspace}
\newcommand{\webpagefull}{\href{http://www.robots.ox.ac.uk/~vgg/research/avobjects}{http://www.robots.ox.ac.uk/$\sim$vgg/research/avobjects}\xspace}

\begin{document}
\pagestyle{headings}
\mainmatter
\def\ECCVSubNumber{2957}  

\title{Self-Supervised Learning of \\Audio-Visual Objects from Video}

\titlerunning{Self-Supervised Learning of Audio-Visual Objects from Video}
%
\author{Triantafyllos Afouras$^{1}$ \and Andrew Owens$^{2}$ \and \\ 
Joon Son Chung$^{1,3}$ \and Andrew Zisserman$^{1}$}
\authorrunning{T. Afouras et al.}
\institute{$^{1}$University of Oxford,
$^{2}$University of Michigan, 
$^{3}$Naver Corporation \\
}
\maketitle

\begin{abstract}
Our objective is to transform a video into a set of discrete audio-visual objects using self-supervised learning.
To this end, we introduce a model 
that uses attention to localize and group sound sources, 
 and optical flow to aggregate information over time.
We demonstrate the
effectiveness of the audio-visual object embeddings that our model learns by using them for four 
downstream speech-oriented tasks: 
(a) multi-speaker sound source separation, (b) localizing and tracking speakers,  (c) correcting misaligned audio-visual data, and (d) active speaker detection. Using our representation, these tasks can be solved entirely by training on unlabeled video, without the aid of object detectors.
We also demonstrate the generality of our method by applying it to non-human speakers, including
cartoons and puppets.
Our model significantly outperforms other self-supervised approaches, and
obtains performance competitive with methods that use supervised face detection. 
\end{abstract}

\vspace{-4mm}
\begin{figure}
\includegraphics[width=\textwidth]{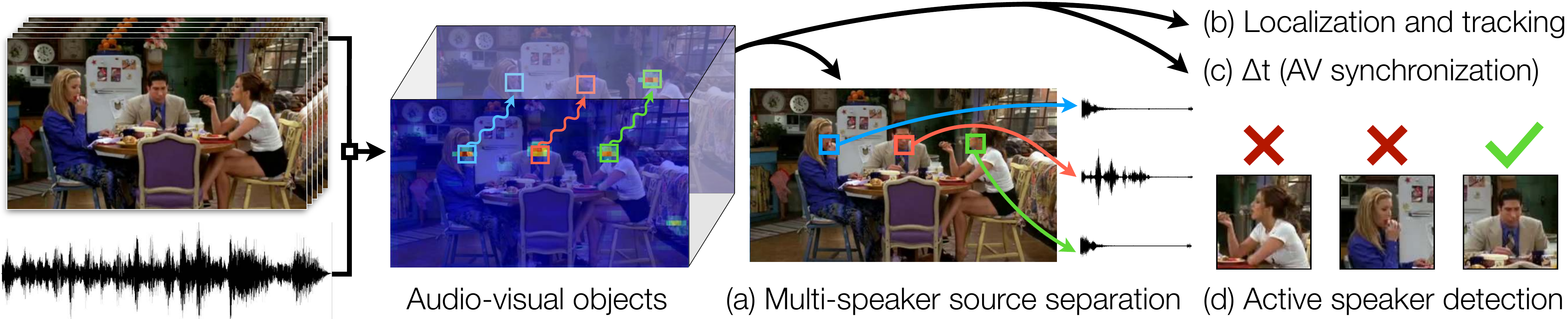}
\vspace{-5mm}
\caption{
We learn through self-supervision to represent a video as a set of discrete {\em audio-visual objects}. 
Our model groups a scene into object instances and represents each one with a feature embedding.
We use these embeddings for speech-oriented tasks that typically require object detectors: (a) multi-speaker source separation, (b) speaker localization, (c) synchronizing misaligned audio and video, and (d) active speaker detection.
Using our representation, these tasks can be solved without any labeled data, and on domains where off-the-shelf detectors are  not available, such as cartoons and puppets.
Please see our webpage for videos: \webpagefull.}

\label{fig:teaser}
\end{figure}
\vspace{-3mm}

\vspace{-4mm}
\section{Introduction}

When humans organize the visual world into objects, hearing provides cues that affect the perceptual
grouping process. We group different image regions together not only because they look alike, or
move together, but also because grouping them together helps us explain the {\em causes} of
co-occurring audio signals. 

In this paper, our objective is to replicate this organizational capability, by designing a model that can ingest raw
video and transform it into a set of  {\em discrete audio-visual objects}. The network is trained using only self-supervised learning
from audio-visual cues. 
We demonstrate this capability on videos containing talking heads.

This organizational task must overcome a number of challenges if it
is to be applicable to raw videos in the wild: (i) there are potentially many visually similar sound generating objects in the scene 
(multiple heads in our case), and the model must correctly attribute the sound to the actual sound source; (ii) these objects may move over time; and (iii) there can be multiple other objects in the scene (clutter) as well.

To address these challenges, we build upon recent works on self-supervised audio-visual localization.
These include video methods that find motions temporally synchronized with
audio onsets~\cite{Chung16a,Owens2018b,korbar18}, and single-frame
methods~\cite{senocak2018learning,Arandjelovic18objects,Owens2018a,harwath2018jointly}
that find regions that are likely to co-occur with the 
audio.  
However, their output is a typically a ``heat map'' that indicates whether a given pixel is likely (or
unlikely) to be attributed to the audio; they 
do not group a scene into {\em discrete objects};
and, if only using semantic correspondence, then they cannot
distinguish which, of several, object instances is making a sound. 

Our first contribution is to propose a network that addresses all three of these challenges; it is able to use synchronization cues to detect sound
sources, group them into distinct instances, and track them over
time as they move.
Our second contribution is to demonstrate that object embeddings obtained from this network facilitate
a number of audio-visual downstream tasks that have previously required  hand-engineered supervised pipelines.

As illustrated in \sfig{fig:teaser}, we demonstrate that the embeddings enable: (a)
multi-speaker sound source separation~\cite{Afouras18,ephrat2018looking}; (b) detecting and tracking talking heads; (c) aligning misaligned recordings~\cite{Chung18a,chung2019perfect}; and (d) detecting active
speakers, i.e.\ identifying which speaker is talking~\cite{roth2019ava,Chung16a}. 
In each case, we significantly outperform other self-supervised
localization methods, and obtain comparable (and in some cases better) performance to prior methods that are trained using stronger supervision, despite the fact that  we learn to perform them entirely
from a raw audio-visual signal.

The trained model, which we call the Look Who's Talking Network (\modelname), is essentially ``plug and play" in that,
once trained on unlabeled data (without preprocessing), it can be applied directly to other video material.
It can easily be fine-tuned for other audio-visual
domains: we demonstrate this functionality on active speaker detection for 
non-human speakers, such as animated characters in {\em The Simpsons} and puppets in {\em Sesame Street}. This demonstrates
the generality of the model and learning framework, since this is a domain where off-the-shelf supervised methods, such as methods that use
face detectors,  cannot transfer without additional labeling.

\mysect{Related work}

\vspace{3mm}

\xpar{Sound source localization.} 
Our task is closely related to the {\em sound source localization}
problem, i.e.\ finding the location in a video that is the source of a
sound.
Early work performed localization~\cite{hershey1999audio,fisher2000learning,kidron2005pixels,Barzelay07} and segmentation~\cite{izadinia2012multimodal} by doing inference on simple probabilistic models, such as methods based on canonical correlation analysis.

Recent efforts learn audio and video representations using self-supervised
learning~\cite{Owens2018b,Chung16a,korbar18} with {\em synchronization} as the proxy task: the network has to
predict whether video and audio are temporally aligned (or
synthetically shifted).
Owens and Efros~\cite{Owens2018b} show via heat-map visualizations
that their network often attends to sound sources, but do not quantitatively evaluate their model. Recent
work~\cite{khosravan2018attention} added an attention mechanism to
this model.  Other work has detected sound-making objects 
using {\em correspondence}
cues~\cite{senocak2018learning,Arandjelovic18objects,Owens2018a,tian2018audio,harwath2018jointly,Hu_2019_CVPR,Ramaswamy_2020_WACV,Hu2020CurriculumAL},
e.g. by training a model to predict whether audio and a single video frame come from the same (or different) videos.
Since these models do not use motion and are
trained only to find the correspondence between object appearance  and  sound, 
they would not be able to identify  which of several objects of the same category is the actual source of
a sound. In contrast, our goal is to obtain discrete audio-visual objects from a
scene, even when they bellong to the same category (e.g. multiple talking heads).
In a related line of work, \cite{gan2019self} distill visual object detectors into an audio model
using stereo sound, while \cite{gao2019visual} use spatial information in a scene to convert mono sound to stereo.

\xpar{Active speaker detection (ASD).} Early work on active speaker detection trained simple classifiers on hand-crafted feature sets~\cite{cutler2000look}. Later, Chung and Zisserman~\cite{Chung16a} used synchronization cues to solve the active speaker detection problem. They used a hand-engineered face detection and tracking pipeline to select candidate speakers, and ran their model only on cropped faces. In contrast, our model learns to do ASD entirely from unlabeled data. 
Chung \etal \cite{chung2019said} extended the pipeline by enrolling speaker models from visible speaking segments. Recently, Roth~\etal~\cite{roth2019ava} proposed an active speaker detection dataset and evaluated a variety of supervised methods for it.

\xpar{Source separation.} In recent years, researchers have proposed a variety of methods for
separating the voices of multiple speakers in a
scene~\cite{ephrat2018looking,Afouras18,Owens2018b,gabbay2018seeing}. These methods either only
handle a single on-screen speaker~\cite{Owens2018b} or use hand-engineered, supervised face
detection pipelines. Afouras \etal~\cite{Afouras18} and Ephrat \etal~\cite{ephrat2018looking}, for
example, detect and track faces and extract visual representations using off-the-shelf packages. In
contrast, we use our model to separate multiple speakers entirely via self-supervision.  

Other recent work has explored separating the sounds of musical
instruments and other sound-making objects.  Gao~\etal~\cite{Gao18graum,gao2019co} use semantic object
detectors trained on instrument categories, while~\cite{zhao2018sound,rouditchenko2019self} do not explicitly
group a scene into objects and instead either pool the visual features or produce a per-pixel map that
associates each pixel with a separated audio source. 
Recently,~\cite{zhao2019sound} added motion information from  optical flow. We, too,
use flow in our model, but instead of using it as a {\em cue} for
motion, we use it to integrate information from moving objects over
time~\cite{Pfister15a,Gadde2017SemanticVC} in order to track them.
In concurrent work ~\cite{Hu2020CurriculumAL} propose a model that groups and
separates sound sources by clustering audio and video embeddings.

\xpar{Representation learning.} 
In recent years, researchers have proposed a variety of self-supervised learning methods for
learning representations from
images~\cite{Doersch15a,Wang15,oord2018representation,tian2019contrastive,misra2020pirl,he2019moco,henaff2019data,chen2020simple},
videos~\cite{Han19,han2020memdpc} and
multimodal data~\cite{owens2016visually,Owens2018a,Arandjelovic17,korbar18,nagrani2020disentangled}.
Often the representation learned by these methods is a feature set (e.g., CNN weights) that
can be adapted to downstream tasks by fine-tuning. 
By contrast, we learn an additional {\em attention mechanism} 
that can be used to group discrete objects of interest for downstream speech tasks. 

\mysect{From unlabeled video to audio-visual objects}
\begin{figure*}[t]
\centering
\includegraphics[width=\textwidth]{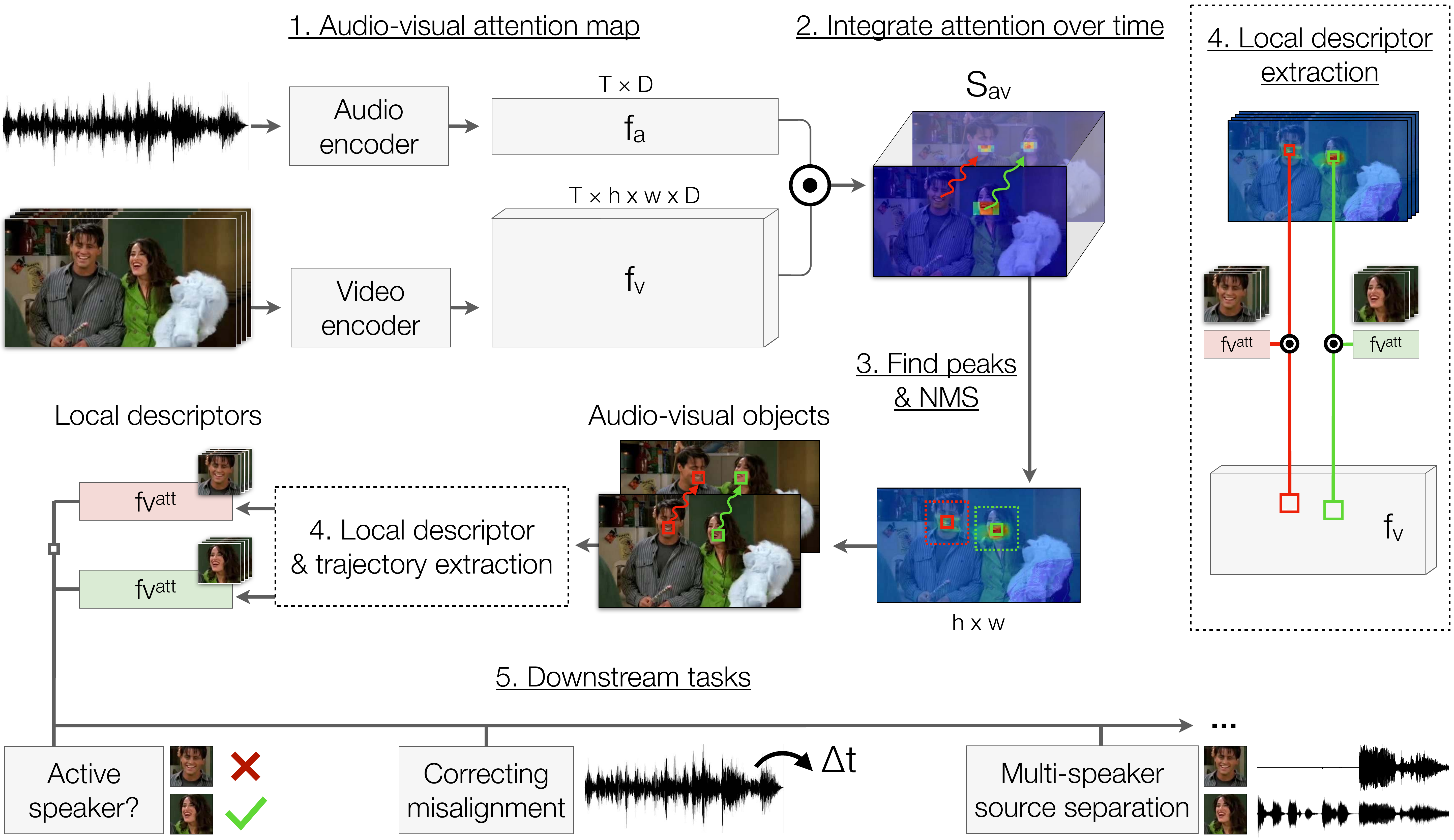}

  \caption{ 
{\bf The Look Who's Talking Network (\modelname)}:
(1) Computes an audio-visual attention map $S_{av}$ by solving a synchronization task,
(2) accumulates attention over time, 
(3) selects {\em audio-visual objects} by computing the $N$ highest peaks with non-maximum suppression (NMS) from the accumulated attention map, each
corresponding to a trajectory of the pixel over time;
(4) for every audio-visual object, it extracts embedding vectors from a spatial window $\rho$, using the local attention map $S_{av}$ to select visual features, and
(5) provides the audio-visual objects as inputs to downstream tasks.
}
\label{fig:architecture}
\vspace{-5mm}

\end{figure*}

Given a video, the function of our model is to detect and track (possibly several) audio-visual objects, and extract embeddings for each of them. 
We represent an audio-visual object as the trajectory of a potential sound source through space and
time, which in the domain that we experiment on is often the track of a ``talking head''. Having obtained these trajectories, we use them to extract embeddings 
that can be then used for downstream tasks.

In more detail, our model uses a bottom-up grouping procedure to
propose discrete audio-visual objects from raw video. It first
estimates local (per-pixel and per-frame) synchronization evidence,
using a network design that is more fine-grained in space and time
than prior models. It then aggregates this evidence over time via
optical flow, thereby allowing the model to obtain robustness to
motions, and groups the aggregated attention into sound sources by
detecting local maxima. The model represents each object
as a separate embedding, temporal track, and attention map
that can be adjusted in downstream tasks. 

We will now give an overview of the model, which is shown in Figure~\ref{fig:architecture}, followed by the learning framework
which uses self-supervision based on synchronization.
For architecture details refer to Appendix~\ref{app:architecture_details}.

\mysubsect{Estimating audio-visual attention}

Before we group a scene into sound sources, we estimate a per-pixel
attention map that picks out the regions of a video whose motions have
a high degree of synchronization with the audio. 
We propose an attention mechanism that provides highly localized spatio-temporal attention, and
which is sensitive to speaker motion.  As in~\cite{Arandjelovic18objects,harwath2018jointly}, we
estimate audio-visual attention via a multimodal embedding (Figure~\ref{fig:architecture}, step~1).
We learn vector embeddings for each audio clip and embedding vectors for each 
pixel, such that if a pixel's vector has a high dot product with that of the audio, then it is
likely to belong to that sound source. For this, we use a two-stream architecture similar to those in other sound-source localization work
~\cite{Arandjelovic18objects,harwath2018jointly,senocak2018learning}, with a network backbone similar to~\cite{chung2019said}. 
We now describe this model in more detail.

\xpar{Video encoder.} Our video feature encoder is a spatio-temporal VGG-M~\cite{chatfield2014return} with a 3D convolutional layer first, followed by a stack of 2D
convolutions. 
Given a \mbox{$T \times H \times W \times 3$} input RGB video, it extracts a video embedding map
$f_v(x,y,t)$ with dimensions 
\mbox{$T \times h \times w \times D$}.

\xpar{Audio encoder.}
The audio encoder is a VGG-M network operating on log-mel spectrograms, treated as single-channel images.
Given an audio segment, it extracts a $D$-dimensional embedding $f_a(t)$ for
every corresponding video frame~$t$.

\xpar{Computing fine-grained attention maps.}
For each space-time pixel, we ask: how correlated is it with the events in the audio?
To estimate this, we measure the similarity between the audio and visual features
at every spatial location.
For every space-time feature vector $f_v(x, y, t)$, we compute the cosine similarity with the audio feature vector $f_a(t)$: 
\begin{equation}
  {S}_{av}(x,y,t) = f_v(x,y,t) {\cdot}  f_a(t),
  \label{eq:av_score}
\end{equation}
where we first $l_2$ normalize both features. We refer to the result, ${S}_{av}(x,y,t)$, as the \textit{audio-visual attention map}. 

\mysubsect{Extracting audio-visual objects} \label{sec:extractingheads}

Given the audio-visual evidence, we parse a video into object representations. 

\xpar{Integrating evidence over time.} Audio-visual objects may only intermittently make sounds. Therefore, we need to integrate sparse attention evidence over time. We also need to group and track sound sources {\em between} frames, while accounting for camera and object motion. 
To make our model more robust
to these motions, we aggregate information over time using optical flow (Figure~\ref{fig:architecture}, step 2). We extract dense optical flow
for every frame, chain the flow values together to obtain long-range tracks, and average the attention scores over these tracks.
Specifically, if $\mathcal{T}( x, y, t)$ is the tracked location of pixel $(x, y)$ from frame $1$
to the later frame $t$, we compute the score:
\begin{equation} 
  S_{av}^{tr} (x,y) = \frac{1}{T} \sum_{t = 1}^{T} S_{av}(\mathcal{T}( x, y, t), t),
\end{equation}
where we perform the sampling using bilinear interpolation. 
The result is a 2D map containing a score for the future trajectory of every pixel of the initial
frame through time. Note that any tracking method can be used in place of optical flow (e.g. with explicit occlusion handling); we use optical flow for simplicity.

\begin{figure}[t]
\centering 
\includegraphics[width=\linewidth]{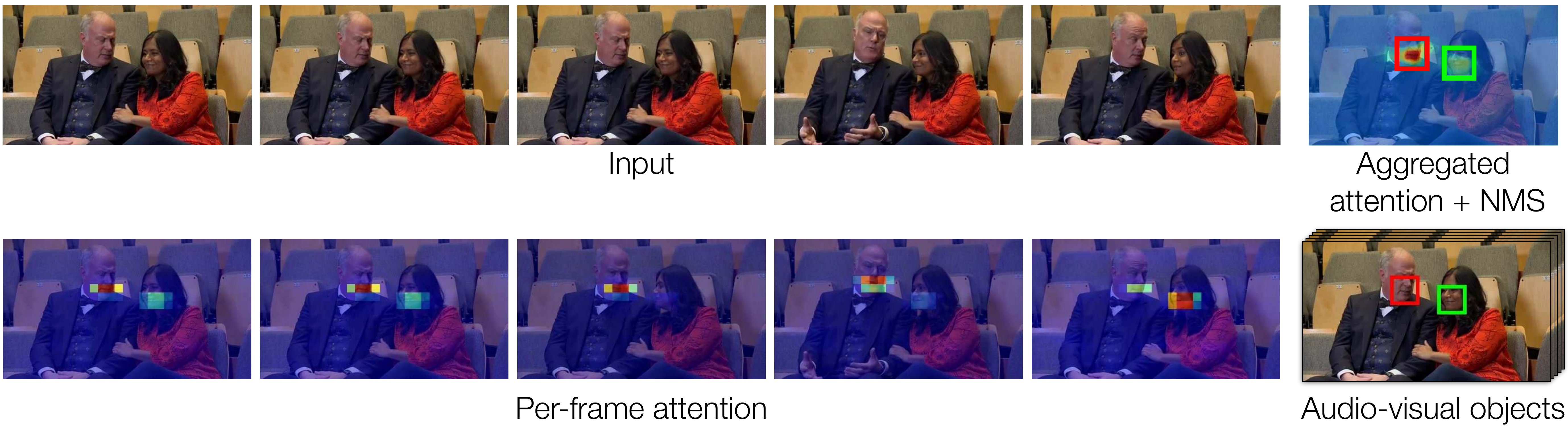}\vspace{-2mm}

\caption{{\bf Intermediate representations from our model}. We show the per-frame attention maps
  ${S}_{av}(t)$, the aggregated attention map
${S}_{av}^{tr}$ and the two highest scoring extracted audio-visual objects.
We show the audio-visual objects for a single frame, with a square of constant width.
} \figvspace

\label{fig:pipeline_examples} 
\end{figure}

\xpar{Grouping a scene into instances.}
To obtain discrete audio-visual objects, we detect spatial local maxima (peaks) on the temporally aggregated synchronization maps, and apply non-maximum
suppression (NMS). More specifically, we find peaks in the time-averaged synchronization map,
$S_{av}^{tr}(x, y)$, and sort them in decreasing order; we then choose the peaks greedily, each time
suppressing the ones that are within a $\rho \times \rho$ box.
The selected peaks can be now viewed as distinct audio-visual objects. Examples of the intermediate representations extracted at the steps described so far are shown in Figure~\ref{fig:pipeline_examples}.

\xpar{Extracting object embeddings.} 
Now that the sound sources have been grouped into distinct audio-visual objects, we can extract feature embeddings for each
one of them that we can use in downstream tasks. Before extracting these features, we locate the position of the sound source in each frame. A simple strategy for this would be to follow the object's optical flow track throughout the video. However, these tracks are imprecise and may not correspond precisely to the location of the sound source. Therefore, we ``snap" to the track location to the nearest peak in the attention map. 
More specifically, in frame $t$, we search in an area of $\rho \times \rho$ centered on the tracked location $\mathcal{T}( x, y, t)$,
and select the pixel location with largest attention value. 
Then, having tracked the sound source in each frame, we select the corresponding spatial feature vector from the visual feature map $f_v$ (Figure~\ref{fig:architecture}, step 4). These per-frame embedding features, $f_{v}^{att}(t)$, can then be used to solve downstream tasks (Section~\ref{sec:downstream}). One can equivalently view this procedure as an audio-visual attention mechanism that operates on $f_v$.

\mysubsect{Learning the attention map} \label{sec:attention}

Training our model amounts to learning the attention map $S_{av}$ on which the audio-visual objects are subsequently extracted. We obtain this map
by solving a self-supervised audio-visual synchronization task~\cite{Owens2018b,korbar18,Chung16a}:
we encourage the embedding at each pixel to be correlated with the true audio and uncorrelated with
shifted versions of it. We estimate the synchronization evidence for each frame by aggregating the per-pixel synchronization scores. Following common practice in multiple instance learning~\cite{Arandjelovic18objects}, we measure the per-frame evidence by the maximum spatial response:
\begin{equation} 
  S_{av}^{att}(t) =  \max_{x,y} S_{av}(x,y,t). \label{eq:framepool}
\end{equation}

We maximize the similarity between a video frame's true audio
track while minimizing that of $N$ shifted (i.e. misaligned) versions of the audio.
Given visual features $f_v$ and true audio $a_i$, we sample $N$ other audio segments from the same video clip: $a_1, a_2, ..., a_N$, and minimize the contrastive loss~\cite{oord2018representation,chung2019perfect}:
\begin{equation}
\small
    \mathcal{L} = -\log \frac{\exp(S_{av}^{att}(v,a_i))}{\exp(S_{av}^{att}(v,a_i)) + \sum_{j=1}^N \exp(S_{av}^{att}(v,a_j))}.
\end{equation}
For the negative examples, we select all audio features (except for the true example) in a temporal window centered on the video frame. 

In addition to the synchronization task, we also consider the {\em correspondence} task of Arandjelovi{\'c} and Zisserman~\cite{Arandjelovic18objects}, which chooses negatives audio samples from random video clips. Since this problem can be solved with even a single frame, it results in a model that is less sensitive to motion.

\mysect{Applications of audio-visual object embeddings} \label{sec:downstream}

Having grouped the video into audio-visual objects, we can use the learned representation to perform a variety of tasks that, in previous work, often required face detection: 1) speaker localization, 2) audio-visual
synchronization, 3) active speaker detection, and 4) audio-visual multi-speaker source separation. We also show the generality of our method by applying it to non-human speakers, such as puppets and animated characters. 

\mysubsect{Audio-visual object detection and tracking}

We can use our model for spatially localizing speakers. To do this, we use the tracked location of an audio-visual object in each frame.

\mysubsect{Active speaker detection}
\label{sec:asd}

For every frame in our video, our model can locate potential speakers and decide whether or not they are speaking.
In our setting, this can be viewed as deciding whether an audio-visual object has strong evidence of synchronization in a given frame. 
For every tracked audio-visual object, we extract the visual features $f_v^{att}(t)$ (Sec. \ref{sec:extractingheads}) for each frame $t$. We then obtain a score that indicates how strong the audio-visual
correlation for frame $t$ is, by computing the dot product: $f_v^{att}(t){\cdot}f_a(t)$.
Following previous work~\cite{Chung16a}, we threshold the result to make a binary decision (active speaker or not). 

\mysubsect{Multi-speaker source separation}
\label{subsec:multispk}

\begin{figure}[t]
\centering 
\includegraphics[width=\columnwidth]{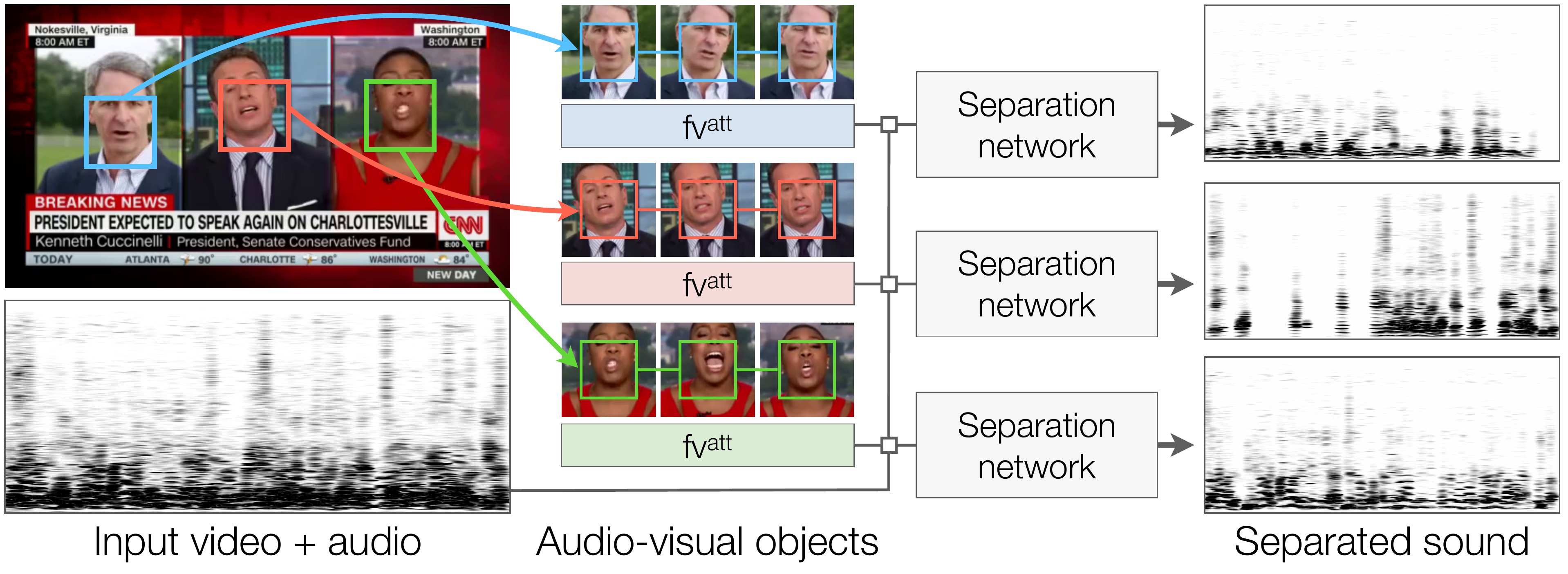}
\vspace{-4mm}
\caption{{\bf Multi-speaker separation}. We isolate the sound of each speaker's voice by
combining our audio-visual objects with a network similar to~\cite{Afouras18}. Given a spectrogram of a noisy sound mixture, the network isolates the voice of each speaker, using the visual features provided by their audio-visual object.} \vspace{-5mm}
\label{fig:sep} 
\end{figure}

Our audio-visual objects can also be used for separating the voices of speakers in a video.  We consider the {\em multi-speaker} separation problem~\cite{Afouras18,ephrat2018looking}: given a video with multiple people speaking on-screen (\eg, a television debate show), we isolate the sound of each speaker's voice from the audio stream. We note that this problem is distinct from on/off-screen audio separation ~\cite{Owens2018b}, which requires only a single speaker to be on-screen.

We train an additional network that, given a waveform containing an audio mixture and an
audio-visual object, isolates the speaker's voice (Figure~\ref{fig:sep}, full details
in Appendix~\ref{app:architecture_details}).
We use an architecture that is similar to~\cite{Afouras18}, but conditions on our self-supervised
representations instead of detections from a face detector. More specifically, the method of~\cite{Afouras18} runs a face detection and tracking system on a video, computes CNN features on each crop, and then feeds those to a source separation network. We, instead, simply provide the same separation network with the embedding features $f_v^{att}(t)$. 


\mysubsect{Correcting audio-visual misalignment}
We can also use our model to correct misaligned audio-visual data --- a problem that often occurs in the recording and television broadcast process. We follow the problem formulation proposed by Chung and Zisserman~\cite{Chung16a}. While this is a problem that is typically solved using supervised face detection~\cite{Chung16a,chung2019perfect}, we instead tackle it with our learned model. During inference, we are given a video with unsynchronized audio and video tracks, and we shift the audio to discover the offset
$\hat{\Delta t}$ that maximizes the audio-visual evidence:  
\begin{equation}
  \small
  \hat {\Delta t} = \argmax_{\Delta t} \frac{1}{T} \sum_{t=1}^T S_{{\Delta t}}^{att}(t),
\end{equation}
where $S_{{\Delta t}}^{att}(t)$ is the synchronization score of frame $t$ after shifting the audio by ${\Delta t}$. Note that this can be estimated efficiently by simply recomputing the dot products in Equation~\ref{eq:av_score}.

In addition to treating this alignment procedure as a stand-alone application, we also use it as a
preprocessing step for our other applications (a common practice in other speech analysis
work~\cite{Afouras18}). When given a test video, we first compute the optimal offset $\hat {\Delta
t}$, and use it to shift the audio accordingly. We then recompute $S_{av}(t)$ from the synchronized embeddings.

\mysect{Experiments}
\subsection{Datasets} \label{sec:datasets}
\vspace{1.5mm}
\xpar{Human speech.}
We evaluate our model on the Lip Reading Sentences (LRS2 and LRS3) datasets
and the Columbia active speaker dataset.
LRS2~\cite{Afouras19} and LRS3~\cite{Afouras18d} are audio-visual speech datasets containing 224
and 475 hours of videos respectively, along with ground truth face tracks of the speakers. 
The Columbia dataset~\cite{Chakravarty16} contains footage from an 86-minute panel discussion,
where multiple individuals take turns in speaking, and contains approximate bounding boxes and active speaker labels, {\em i.e.} whether a visible face is speaking at a given point in time.
All datasets provide (pseudo-)ground truth bounding boxes obtained via face detection, which we use for evaluation. 

We resample all videos to a resolution of $H \times W = 270\times480$ pixels before feeding them to our model, which outputs $h\times w = 18\times31$ attention maps. 
We train and evaluate all models (except for those with non-human speakers) on LRS2, and use LRS3 only for evaluation.

\xpar{Non-human speakers}
To evaluate our method on non-human speakers, we collected
television footage from {\em The Simpsons} and
{\em Sesame Street} shows (Table~\ref{tab:datasets}).
We trained on the raw footage without performing any preprocessing, except for splitting the videos into scenes.
For testing, we collected ASD and speaker localization labels, using the VIA tool~\cite{dutta2019vgg}:
we asked human annotators to label frames that they believed to contain an active speaker and to localize them.
Videos were annotated sparsely with only a few frames per video clip.
For every dataset, we create two test sets.
In the \emph{single-head} set, the clips are constrained to contain a single active speaker,
with no other faces.
The second test subset, \emph{multi-head}, may contain multiple heads --- talking or not --- and also
a variety of cases with no relevant speech (non-talking heads, background, title sequences etc.). 
We summarize the statistics of the test sets in Table~\ref{tab:datasets}. 
For full details refer to Appendix~\ref{app:non-human}.

\mysubsect{Training details}
\label{subsec:trainingdetails}

\begin{figure*}[t]
 \centering 
 \vspace{5pt}

\includegraphics[width=0.162\columnwidth]{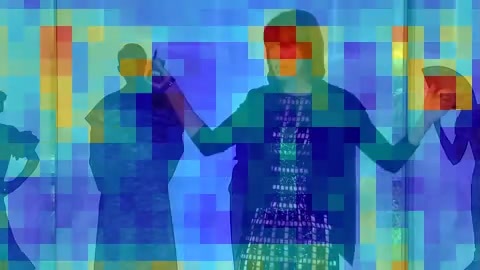}\hspace{.3mm}\includegraphics[width=0.162\columnwidth]{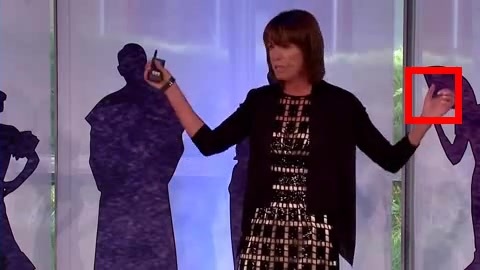}\hspace{0.95mm}\includegraphics[width=0.162\columnwidth]{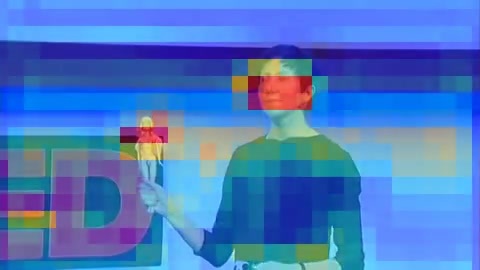}\hspace{.3mm}\includegraphics[width=0.162\columnwidth]{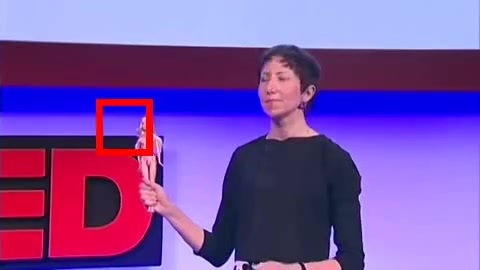}\hspace{0.95mm}\includegraphics[width=0.162\columnwidth]{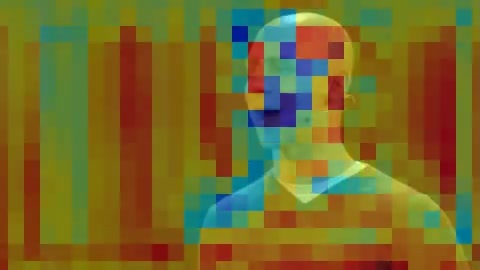}\hspace{.3mm}\includegraphics[width=0.162\columnwidth]{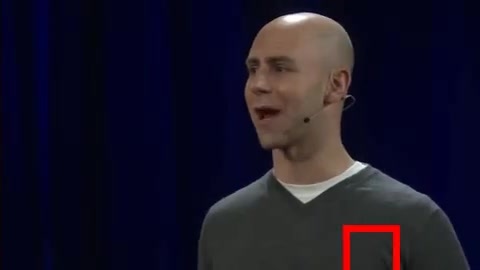}\hspace{0.95mm} 
AVOL-net~\cite{Arandjelovic18objects}\\
\includegraphics[width=0.162\columnwidth]{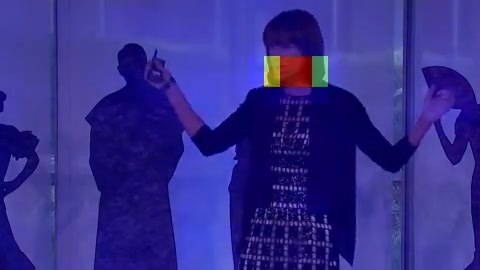}\hspace{.3mm}\includegraphics[width=0.162\columnwidth]{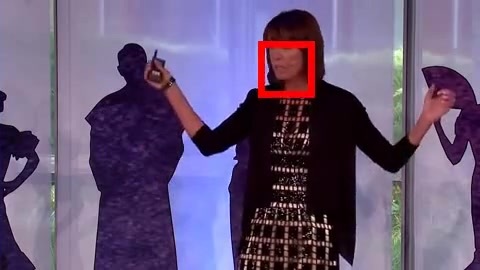}\hspace{0.95mm}\includegraphics[width=0.162\columnwidth]{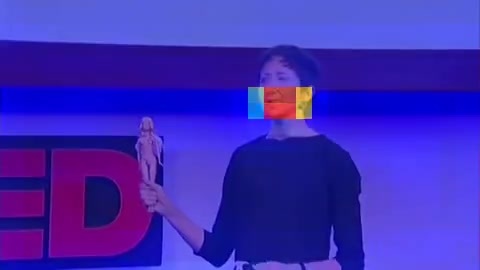}\hspace{.3mm}\includegraphics[width=0.162\columnwidth]{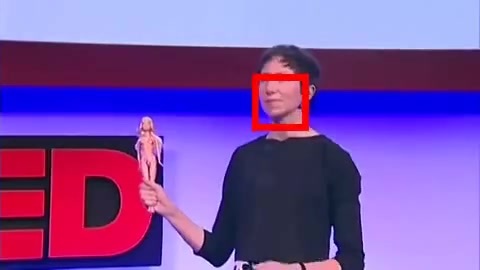}\hspace{0.95mm}\includegraphics[width=0.162\columnwidth]{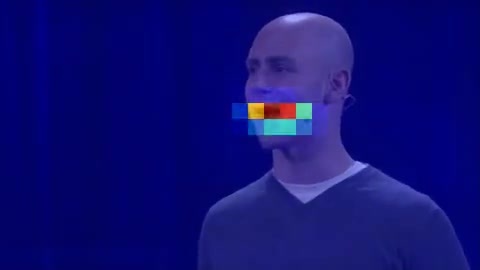}\hspace{.3mm}\includegraphics[width=0.162\columnwidth]{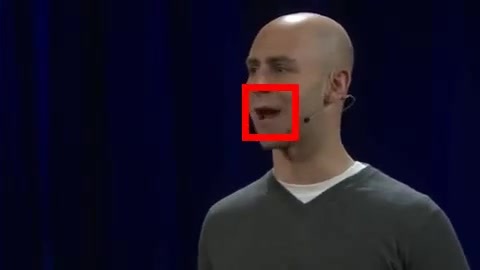}\hspace{0.95mm}
Our model

\vspace{-1.5mm}
 \caption{{\bf Talking head detection and tracking on LRS3 datasets.} 
 For each of the 4 examples, we show the audio-visual attention score on every spatial location
 for the depicted frame, and a bounding box centered on the largest value, indicating the speaker location. 
 Please see our \webpage for video results.
 }
 \label{fig:loc_baselines} 
 \end{figure*}

 \begin{figure}
   \centering 
      \includegraphics[width=0.162\columnwidth]{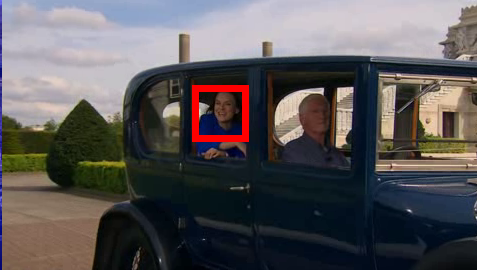}\hspace{.3mm}\includegraphics[width=0.162\columnwidth]{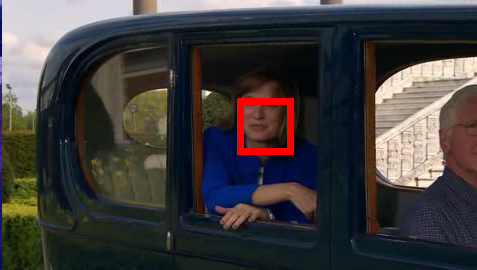}\hspace{.95mm}\includegraphics[width=0.162\columnwidth]{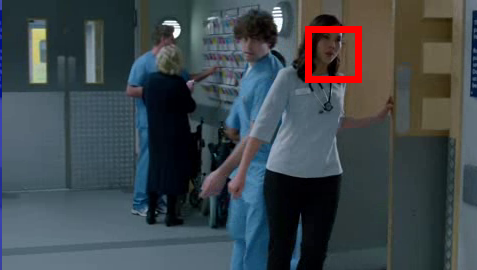}\hspace{.3mm}\includegraphics[width=0.162\columnwidth]{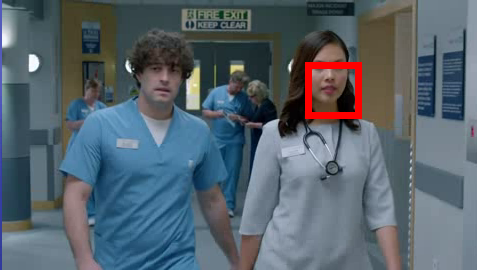}\hspace{.95mm}\includegraphics[width=0.162\columnwidth]{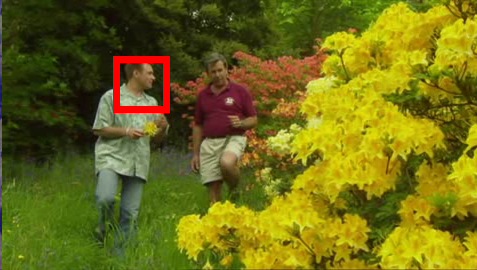}\hspace{.3mm}\includegraphics[width=0.162\columnwidth]{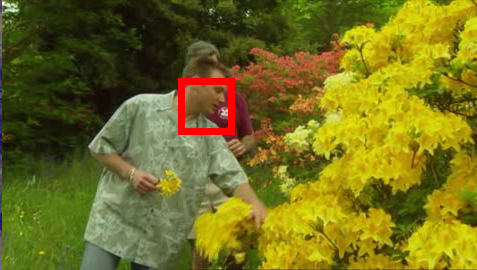}

    \vspace{.6mm}
   \includegraphics[width=0.162\columnwidth]{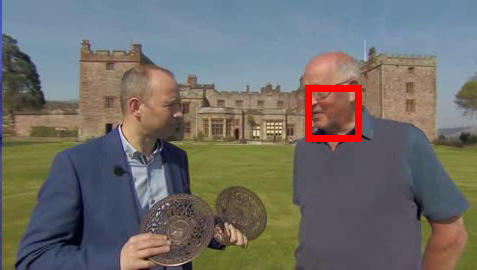}\hspace{.3mm}\includegraphics[width=0.162\columnwidth]{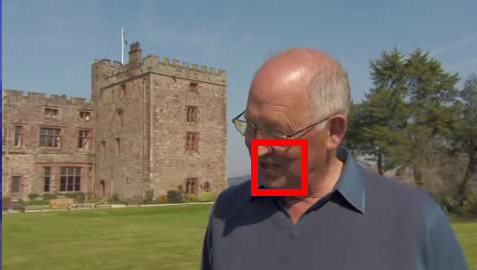}\hspace{.95mm}\includegraphics[width=0.162\columnwidth]{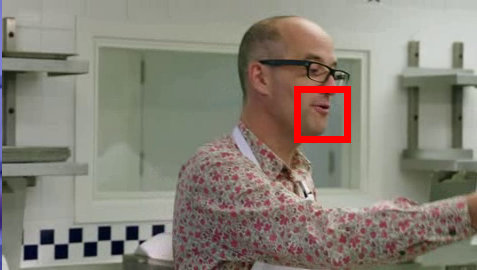}\hspace{.3mm}\includegraphics[width=0.162\columnwidth]{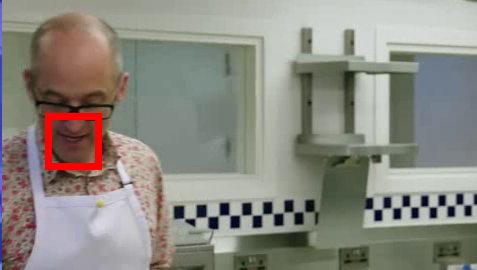}\hspace{.95mm}\includegraphics[width=0.162\columnwidth]{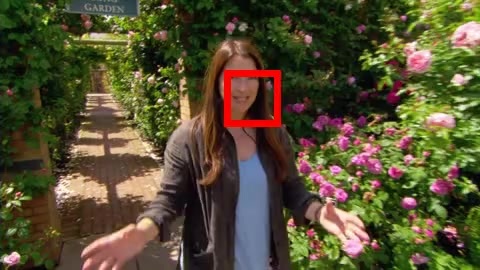}\hspace{.3mm}\includegraphics[width=0.162\columnwidth]{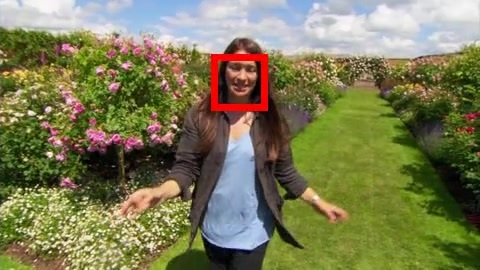}
   
   \vspace{-.9mm}
   \caption{{\small {\bf Handling motion}: Talking head detection and tracking on continuous scenes from the validation set of LRS2. 
       Despite the significant movement of the speakers and the camera, our method 
       accurately tracks them. 
    }}\vspace{0.1mm}
   \label{fig:camera_motion} 
   \vspace{-20pt}

 \end{figure}

\newpara\noindent{\textbf{Audio-visual object detection training.}}
To make training easier, we follow~\cite{korbar18} and use a simple learning
curriculum. At the beginning of training, we sample negatives from random video clips, then switch to shifted audio tracks later in training. 
To speed up training, we also begin by taking the mean dot product (Eq.~\ref{eq:framepool}), and then switch to the maximum. We set $\rho$ to 100 pixels.

\xpar{Source separation training} 
Training takes place in two steps: we first train our model to produce audio-visual objects by solving a synchronization problem. 
Then, we train the multi-speaker separation network on top of these learned 
representations.   
We follow previous work~\cite{Afouras18,ephrat2018looking} and use
a mix-and-separate learning procedure.
We create synthetic videos containing multiple talking speakers by 1) selecting two or three videos at
random from the training set, depending on the experiment, 2) summing their waveforms together, and 3) vertically concatenating
the video frames together. The model is then tasked with extracting a number of talking heads equal
to the number of mixed videos and predicting an original corresponding waveform for each.

\xpar{Non-human model training} 
We fine-tune the best model from LRS2 separately on each of the two datasets with non-human speakers.
The lip motion for non-human speakers, such as the motion of a puppet's mouth, 
is only loosely correlated with speech, suggesting that there is less of an advantage to obtaining our negative examples from temporally shifted audio. We therefore sample our negative audio examples from other video clips rather than from misaligned audio (Section \ref{sec:attention}) when computing attention maps, but keep the rest of the architecture the same.
\mysubsect{Results}

\begin{figure}[t]
\centering 

\includegraphics[width=0.24\columnwidth]{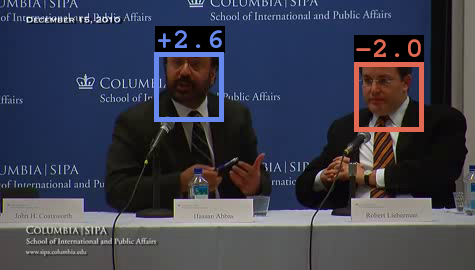}
\includegraphics[width=0.24\columnwidth]{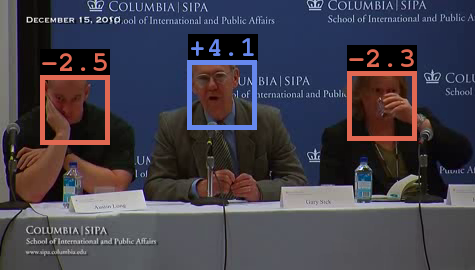}
\includegraphics[width=0.24\columnwidth]{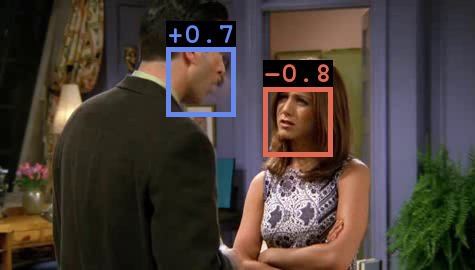}
\includegraphics[width=0.24\columnwidth]{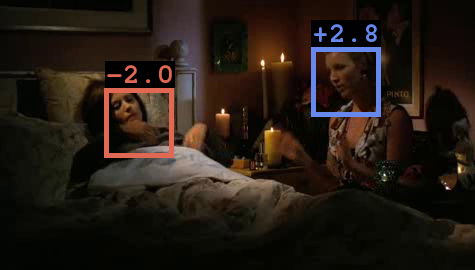}

\vspace{-2mm}
\caption{{\bf Active speaker detection} on the Columbia dataset, and an example from the {\em Friends} TV show.
We show active speakers in \textbf{\textcolor{boxblue}{blue}} and inactive speakers in
\textbf{\textcolor{boxred}{red}}. The corresponding detection scores are noted above the boxes (the
threshold has been subtracted so that positive scores indicate active speakers).
}
\label{fig:asd_col} 
\end{figure}

\newpara\noindent{\textbf{1. Talking head detection and tracking.}}
We evaluate how well our model is able to localize speakers, i.e. talking heads (Table~\ref{tab:face_localization}).
First, we evaluate two simple baselines: the {\em random} one, which selects a
random pixel in each frame and the {\em center} one, which always selects the center pixel.
Next, we compared with two recent sound source localization methods: Owens and Efros~\cite{Owens2018b}
and AVOL-net~\cite{Arandjelovic18objects}. Since these methods require input videos that are
longer than most of the videos in the test set of LRS2, we only evaluate them on LRS3. We also perform several
ablations of our model: To evaluate the benefit of integrating the audio-visual evidence over flow
trajectories, we create a variation of our model called {\em No flow} that, instead, computes the
attention $S_{av}^{tr}$ by globally pooling over time throughout the video. Finally, we also
consider a variation of this model that uses a larger NMS window ($\rho=150$).

\begin{figure}[t]
\centering 
 
\includegraphics[width=0.244\columnwidth]{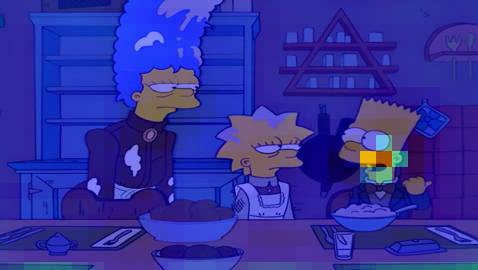}\hspace{.3mm}\includegraphics[width=0.244\columnwidth]{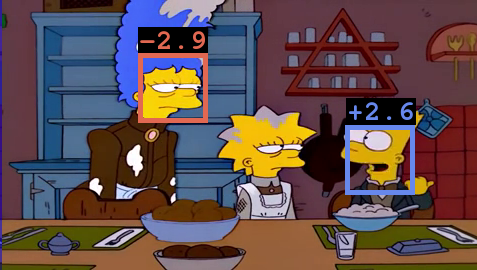}
\hspace{0.95mm}\includegraphics[width=0.244\columnwidth]{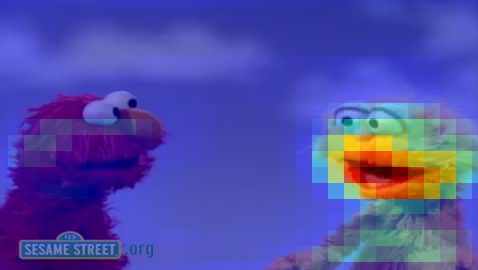}\hspace{.3mm}\includegraphics[width=0.244\columnwidth]{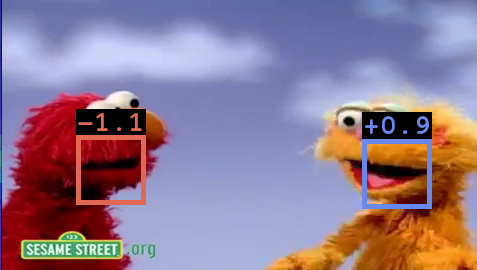}

\includegraphics[width=0.244\columnwidth]{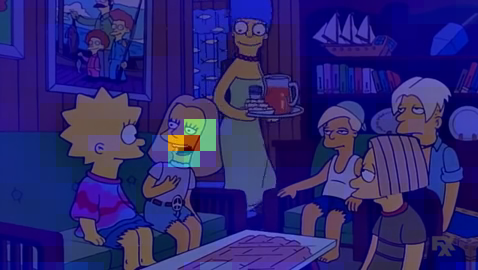}\hspace{.3mm}\includegraphics[width=0.244\columnwidth]{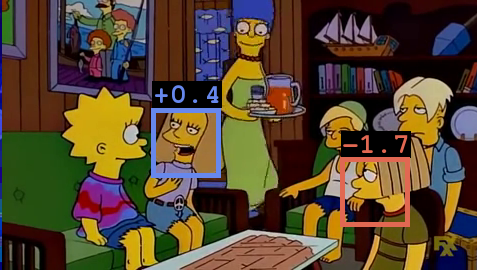}
\hspace{0.95mm}\includegraphics[width=0.244\columnwidth]{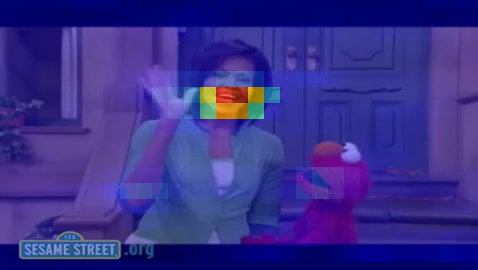}\hspace{.3mm}\includegraphics[width=0.244\columnwidth]{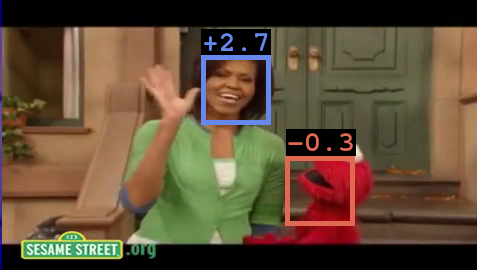}

\includegraphics[width=0.244\columnwidth]{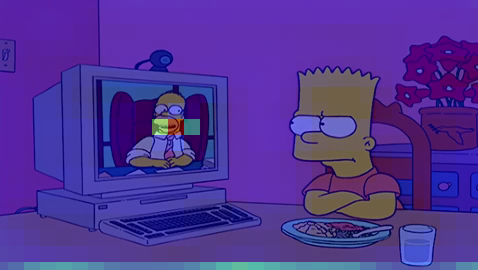}\hspace{.3mm}\includegraphics[width=0.244\columnwidth]{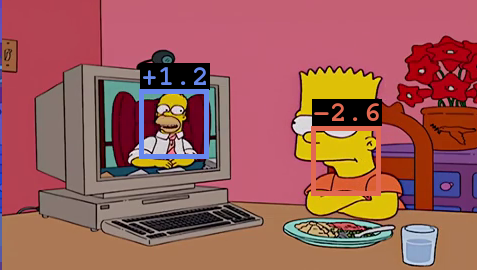}
\hspace{0.95mm}\includegraphics[width=0.244\columnwidth]{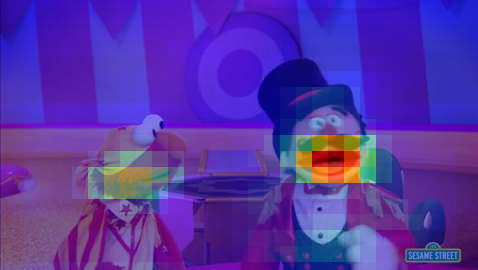}\hspace{.3mm}\includegraphics[width=0.244\columnwidth]{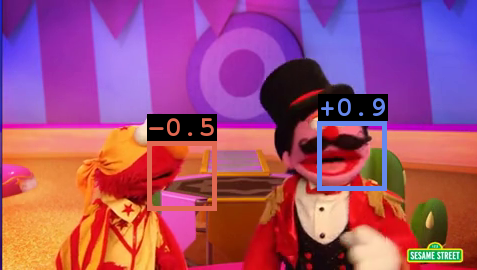}

\vspace{-2mm}
\caption{{\bf Active speaker detection for non-human speakers}. We show the top 2 highest-scoring
  audio-visual objects in each scene, along with the aggregated attention map. Please see our \webpage for video results. 
 } 
\label{fig:cartoons} 
\figvspace
\end{figure}

We found that our method obtains very high accuracy, and that it significantly outperforms
all other methods.
AVOL-net solves a correspondence task that doesn't require motion information, and uses a single
video frame as input. Consequently, it does not take advantage of informative motion, such as moving lips.  As can be seen in Figure~\ref{fig:loc_baselines}, the localization maps produced by AVOL-net~\cite{Arandjelovic18objects} are less precise, as it only loosely associates appearance of a person to speech, and won't consistently focus on the same
region. 
Owens and Efros~\cite{Owens2018b}, by contrast, has a large temporal receptive field, which results in temporally imprecise predictions, causing very large errors when the subjects are moving. 
The {\em No flow} baseline fails to track the talking head well outside the NMS area, and its accuracy is
consequently lower on LRS3. Enlarging the NMS window partially alleviates this issue, but the accuracy is still lower than that of our model.
We note that the LRS2 test set contains very short clips (usually 1-2 seconds long) with predominantly
static speakers, which explains why using flow does not provide an advantage.

We show some challenging examples with significant speaker and camera motion in Figure~\ref{fig:camera_motion}.
Refer to Appendix~\ref{app:motion} for further analysis on robustness to camera and speaker motion.

\captionsetup[subfloat]{labelformat=showtable}

\begin{table}[t] 
\begin{center}
  \subfloat[{\small {\bf Talking head detection and tracking accuracy}. A detection is considered correct if
      it lies within the true bounding box.}]{%
    \scriptsize
    \centering
    \setlength{\tabcolsep}{2.0pt}
    \setlength\extrarowheight{2pt}
    \begin{tabular}[b]{ l r r}
      \toprule
          {\bf Method}   & {\bf LRS2} & {\bf LRS3}  \\ 
          \midrule                              
          Random          &    2.8\%   &   2.9\%        \\  
          Center          &   23.9\%   &  25.9\%        \\  
          Owens \& Efros~\cite{Owens2018b}        &   -   &  24.8\%     \\  
          AVOL-net~\cite{Arandjelovic18objects}        &   -   &    58.1\%    \\ 
          \cdashline{1-3}                              
          No flow &  98.4\%   &  94.2\%        \\  
          No flow + large NMS&   98.8\%   &  97.2\%        \\  
          Full model        &  \bf{99.6}\%   &  \bf{99.7}\%        \\  
          \bottomrule
    \end{tabular}
    \label{tab:face_localization}}~~~
  \subfloat[{\bf Active speaker detection accuracy} on the Columbia dataset~\cite{Chakravarty16}.
      F1 Scores (\%) for each speaker, and the overall average.]{
    \scriptsize
    \centering
    \setlength{\tabcolsep}{1.0pt}
     \setlength\extrarowheight{2.5pt}
\newcolumntype{L}[1]{>{\raggedright\let\newline\\\arraybackslash\hspace{0pt}}m{#1}}
\newcolumntype{C}[1]{>{\centering\let\newline\\\arraybackslash\hspace{0pt}}m{#1}}
\newcolumntype{R}[1]{>{\raggedleft\let\newline\\\arraybackslash\hspace{0pt}}m{#1}}
   \hspace{-5mm}\begin{tabularx}{0.52\linewidth}[b]{ l C{0.055\textwidth} C{0.055\textwidth} C{0.055\textwidth} C{0.055\textwidth} C{0.055\textwidth} | C{0.055\textwidth}}
      \toprule
          {\bf Method}   & \multicolumn{5}{c}{{\bf Speaker}}\\
          & { Bell } & {Boll }  & {Lieb} & {Long} & {Sick} & {Avg.}  \\
          \midrule                              
          Chakravarty~\cite{Chakravarty16}               & 82.9   & 65.8 & 73.6  & 86.9 & 81.8 & 80.2 \\  
          Shahid~\cite{Shahid_2019_ICCV}    & 87.3   & 96.4 & 92.2  & 83.0 & 87.2 & 89.2 \\  
          SyncNet~\cite{Chung16a}                   & 93.7  & 83.4 & 86.8  & 97.7 & 86.1 & 89.5 \\  
          \cdashline{1-7}
          Ours                                     & 92.6   & 82.4 & 88.7  & 94.4 & 95.9 & \textbf{90.8} \\  
          \bottomrule
    \end{tabularx}
    \setlength\extrarowheight{5.3pt}

    \label{tab:asd}
  }
\end{center}
\normalsize
\figvspace
\end{table}
\setcounter{table}{1}

\newpara\noindent{\textbf{2. Active speaker detection.}}
Next, we ask how well our model can determine {\em which} speaker is talking. Following previous work that uses supervised face detection~\cite{Chung16a,Shahid_2019_ICCV}, we
evaluate our method on the Columbia dataset~\cite{Chakravarty16}.
For each video clip, we extract 5 audio-visual objects (an upper bound on the number of speakers),
each of which has an ASD score indicating the likelihood that it is a sound source
(Section~\ref{sec:asd}). We then associate each ground truth bounding box with the audio-visual
object whose trajectory follows it the closest. For comparison with existing work, we report the F1
measure (the standard for this dataset) per individual speaker as well as averaged over all speakers.
For calculating the F1 we set the ASD threshold to the one that yields the Equal Error Rate (EER) for the pretext task on the LRS2 validation set. 
As shown in Table~\ref{tab:asd}, our model outperforms
all previously reported results on this dataset, even though (unlike other methods) it does not use labeled face bounding boxes for training.

\xpar{3. Multi-speaker source separation.}
To evaluate our model on speaker separation, we follow the protocol of ~\cite{Afouras18}. We create synthetic examples from the test set of LRS2, using only videos that are between $2-5$ seconds long, and evaluate performance using Signal-to-Distortion-Ratio (SDR)~\cite{Fevotte05} and Perceptual Evaluation of Speech Quality (PESQ, varies between 0 and 4.5)~\cite{rix2001perceptual} (higher is better for both).
We also assess the intelligibility of the output by computing the Word Error Rate (WER, lower is better)
between the transcriptions obtained with the Google Cloud speech recognition system.
Following~\cite{Afouras18d}, we train and evaluate separate models for 2 and 3 speakers, though
we note that if the number of speakers were unknown, it could be estimated using active speaker detection.

For comparison, we implement the model of Afouras \etal~\cite{Afouras18}, and train it on the same data. For extracting visual features to serve as its input, we use a state-of-the-art audio-visual synchronization model~\cite{chung2019perfect}, rather than the lip-reading features from Afouras \etal~\cite{Afouras19b}. We refer to this model as {\em Conversation-Sync}. 
This model uses bounding boxes from a well-engineered face detection system, and thus represents an approximate upper limit on the performance of our self-supervised model.
Our main model for this experiment is trained end-to-end and uses $\rho=150$. We also performed a
number of ablations: a model that freezes the pretrained audio-visual features and a model with a
smaller $\rho=100$. 

We observed (Table~\ref{tab:sep}) that our self-supervised model obtains results close to those of~\cite{Afouras18}, which is based on supervised face detection. We also asked how much error is introduced by lack of face detection. In this direction we extract the local visual descriptors using tracks obtained with face detectors instead of our audio-visual object tracks. This model, {\em Oracle-BB}, obtains results similar
to ours, suggesting that the quality of our face localization is high.

\captionsetup[subfloat]{labelformat=showtable}

\begin{table}[t] 
\begin{center}
  \subfloat[ {\bf Source separation} on LRS2. \#Spk indicates the number of speakers. The WER on the ground truth signal is 20.0\%.]{
    \scriptsize
    \centering
    \setlength{\tabcolsep}{2.5pt}
    \setlength\extrarowheight{1pt}
\hspace{-10pt}
\begin{tabularx}{0.5\linewidth}{ l @{\hspace{-40pt}} l @{\hspace{2\tabcolsep}} rr @{\hspace{2\tabcolsep}} rr
@{\hspace{2\tabcolsep}} rr } 
 \toprule

 & & \multicolumn{2}{c}{SDR} & \multicolumn{2}{c}{PESQ} &\multicolumn{2}{c}{WER \%} \\  
\addlinespace[2pt]

{\textbf{Method}} $ \backslash $ {\# Spk.} & & \multicolumn{1}{c}{2} & \multicolumn{1}{c}{3}  &
\multicolumn{1}{c}{2} & \multicolumn{1}{c}{3}  & \multicolumn{1}{c}{2} & \multicolumn{1}{c}{3} \\ 
\midrule
Mixed input                           &  & -0.3 & -3.4      & 1.7 & 1.5      & 91.0  & 97.2   \\
Conv.-Sync~\cite{Afouras18}                                           &  &  11.3 & 7.5      & 3.0  & 2.5     & 30.3 & 43.5   \\ 
\midrule
       & Frozen                                       &  10.7 & 7.0      & 3.0  & 2.5      & 30.7 & 44.2   \\ 
Ours   & Oracle-BB                                    &  10.8 & 7.1      & 2.9  & 2.5      & 30.9 & 44.9   \\ 
       & Small-NMS                                   &  10.6 & 6.8      & 3.0  & 2.5      & 31.2 & 44.7   \\
       & Full                                     &  10.8 & 7.2      & 3.0  & 2.6      & 30.4 & 42.0   \\ 
 \bottomrule
\end{tabularx}             
\label{tab:sep}}~~~
    \subfloat[{\bf Audio-visual synchronization} accuracy (\%) evaluation for a given number of input frames.]{
    \scriptsize
    \centering
    \setlength\extrarowheight{7.8pt}
\hspace{-8pt}
\begin{tabular}{ l c c c c c c  } 
 \toprule
 & \multicolumn{6}{c}{{\bf Input frames}} \\
 {\bf Method}   & {5} & {7} & {9} & {11} & {13} & {15}  \\ 
 \midrule
 SyncNet \cite{Chung16a}           &   75.8   &   82.3     &  87.6 & 91.8     & 94.5     & 96.1        \\  
 PM \cite{chung2019perfect}        &   88.1   &   93.8      & 96.4  & 97.9  & 98.7    & 99.1    \\  
 \cdashline{1-7}
 Ours                              &   78.8    &  87.1      & 92.1  &  94.8  & 96.3    &  97.3    \\  
 \bottomrule
\end{tabular}          
    \label{tab:sync} 
  } 
\end{center}
\normalsize
\figvspace
\end{table}
\setcounter{table}{2}


\xpar{4. Correcting misaligned visual and audio data.}
We use the same metric as \cite{chung2019perfect} to evaluate on LRS2. 
The task is to determine the correct audio-to-visual offset within a $\pm$15 frame window. An offset is considered correct if it is within 1 video frame from the ground truth. The distances are averaged over 5 to 15 frames. 
We compare our method to two state-of-the-art synchronization methods: SyncNet~\cite{Chung16a} and the state-of-the-art Perfect Match~\cite{chung2019perfect}. 
We note that \cite{chung2019perfect} represents an approximate upper limit to what we would expect our method to achieve, since we are using a similar network and training objective; the major difference is that we use our audio-visual objects instead of image crops from a face detector. 
The results (Table~\ref{tab:sync}) show that our self-supervised model
obtains comparable accuracy to these supervised methods.

\xpar{5. Generalization to non-human speakers.}%
We evaluate the \modelname model's generalization to non-human
speakers using the {\em Simpsons} and {\em Sesame Street} datasets described in Section~\ref{sec:datasets}.
The results of our evaluation are summarized in Table~\ref{tab:cartoons}. Since supervised speech analysis methods are often based on face detection systems, we compare our method's performance to off-the-shelf face detectors, using the  \emph{single-head} subset.
As a face detector baseline, we use the state-of-the-art RetinaFace~\cite{deng2019retinaface}
detector, with both the MobileNet and ResNet-50 backbones.
We report localization accuracy (as in Table~\ref{tab:face_localization}) and Average Precision (AP).
It is clear that our model outperforms the face detectors in both localization and retrieval
performance for both datasets. 

The second evaluation setting is detecting active speakers in videos from the \emph{multi-head} test set.
As expected, our model's performance decreases in this more challenging scenario;
however, the AP for both datasets indicates that our method can be useful for retrieving the speaker 
in this entirely new domain.
We show qualitative examples of ASD on the {\em multi-head} test sets in \sfig{fig:cartoons}.

\captionsetup[subfloat]{labelformat=showtable}

\begin{table}[t] 
\begin{center}
  \subfloat[{\bf Label statistics} for non-human test sets. S is {\em single head} and M \em{multi-head}. ]{%
    \scriptsize
    \centering
    \setlength{\tabcolsep}{1.0pt}
    \setlength\extrarowheight{2pt}
    \begin{tabular}[b]{ l r r r } 
 \toprule
 {\bf Source}  & {\bf Type}  &  {\bf Clips} &  {\bf Frames}  \\ 
 \midrule                                             
 The Simpsons     & S &  41   &  87    \\  
 The Simpsons     & M &  582 &  251  \\  
 \midrule                                                 
 Sesame Street    & S &  57  &  120  \\  
 Sesame Street    & M &  143  &  424  \\  
 \bottomrule
\end{tabular}             
    \label{tab:datasets}}~~~
  \subfloat[{\bf Non-human speaker evaluation} for ASD and localization tasks on {\em Simpsons} and {\em Sesame Street}. MN: MobileNet;
  RN: ResNet50.]{
  
    \scriptsize
    \centering
    \vspace{-20pt}
    \setlength{\tabcolsep}{2.0pt}
    \setlength\extrarowheight{1pt}
\begin{tabular}[b]{  l rr | rr rr rr r  } 
 \toprule

 & \multicolumn{2}{c}{Loc. Acc} & \multicolumn{6}{c}{ASD AP} \\  
 & \multicolumn{2}{c}{Single-head} & \multicolumn{2}{c}{Single-head} & \multicolumn{2}{c}{Multi-head}  \\  
\addlinespace[2pt]

{\textbf{Method}} & Simp. & Ses.  & Simp. & Ses.  & Simp. & Ses. \\ 
\midrule
       Random                                    &  8.7  & 16.0     &  - & -        &  - & -       \\ 
       Center                                    &  62.0 & 80.1     &  - & -        &  - & -      \\ 
       RetinaFace RN                             &  47.7 & 61.2    &  40.0  & 46.8 &  - & -      \\ 
       RetinaFace MN                             &  72.1 & 70.2     &  60.4  & 52.4      &  - & -      \\ 
\cdashline{1-7}
       Ours                                 &  {\bf 98.8} & {\bf 81.0}   &  {\bf 98.7} & {\bf 72.2}   &  {\bf 85.5} & {\bf 55.6}  \\ 
 \bottomrule

\end{tabular}            
    \label{tab:cartoons}
  } 
\end{center}
\normalsize
\figvspace
\end{table}

\mysect{Conclusion}
In this paper, we have proposed a unified model that learns from raw
video to detect and track speakers. The embeddings learned by the model
are effective for many downstream speech analysis tasks, such as
source separation and active speaker detection, that in
previous work required supervised face detection.

We see our work opening two new directions. The first one is in
extending our object embeddings to other audio-visual speaker tasks, such as
diarizing conversations~\cite{gebru2017audio,chung2019said}, and face/head detection. 
The second one is in
self-supervised representation learning. We have presented a framework that is
well-suited to speech tasks but could also have potential in different domains, such as the analysis of music and ambient sounds. 
For code and models, please see our \webpage.

\vspace{10pt}\noindent\textbf{Acknowledgements.}
We thank V. Kalogeiton for generous help with the annotations and the {\em Friends} videos,
A. A. Efros for helpful discussions, L. Momeni, T. Han and Q. Pleple for proofreading, A. Dutta for
help with VIA, and A. Thandavan for infrastructure support.
This work is funded by the UK EPSRC CDT in AIMS, DARPA Medifor,
and a Google-DeepMind Graduate Scholarship.

\clearpage
%
\bibliographystyle{splncs04}
\bibliography{shortstrings,vgg_local,vgg_other,mybib}

\clearpage
\begin{subappendices}
\renewcommand{\thesection}{\Alph{section}}

\section{Robustness to motion.} \label{app:motion}
To assess the robustness of our method to large camera and speaker motions,
we used optical flow to rank videos by the amount of motion and obtained “high motion” subsets for the
validation set of LRS2 and test set of LRS3.
For LRS2 we used the validation instead of the test set because the videos there are untrimmed and longer.
As shown on Table~\ref{tab:camera_motion} our method maintains good performance even on videos with large camera motion.
The performance drop for our full model on those videos is minimal while the no-flow baseline suffers more.
Further qualitative inspection suggests that camera motion is rarely a source of error.
Please refer to our webpage for examples on these high-motion videos, where the robustness to motion can be observed qualitatively.

\begin{table}[h] 
    \centering
    \caption{ Breakdown of performance on talking head detection for high and
    low motion subsets of LRS2 validation and LRS3 test sets. }
    \label{tab:camera_motion}
    \vspace{5pt}
    \setlength{\tabcolsep}{6.0pt}
    \scriptsize
    \begin{tabular}[b]{ l r r r r} 
      \toprule
                         & \multicolumn{2}{c}{{\bf LRS2-val}}  & \multicolumn{2}{c}{{\bf LRS3-test}}   \\
          {\bf Model}   & \multicolumn{1}{c}{low} & \multicolumn{1}{c}{high} & \multicolumn{1}{c}{low} & \multicolumn{1}{c}{high}      \\ 
          \midrule                              
          No flow      & 97.8\%   & 93.6\%  &  94.8\%      & 88.1\%       \\  
          Full model   & 98.1\%   & 96.1\%  &  99.8\%      & 99.3\%       \\  
          \bottomrule
    \end{tabular}
    \normalsize
\end{table}

\section{Sensitivity to NMS scale}
\begin{figure}[bh!]
 \centering 
 \includegraphics[width=0.5\columnwidth]{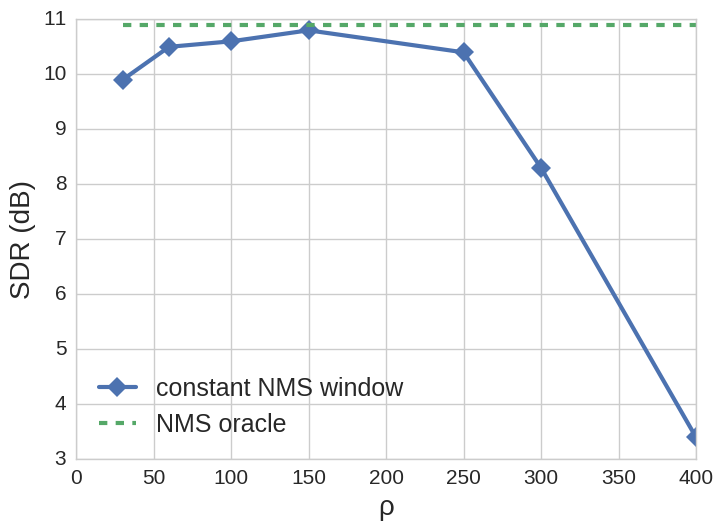}
 \vspace{-.9mm}
 \caption{Source separation performance on LRS2, when varying the NMS window $\rho$.}
 \vspace{0.1mm}
 \label{fig:nms_oracle} 
 \vspace{-20pt}
\end{figure}

Our method uses a constant scale for simplicity, since we found that performance is not very sensitive to it.
To determine the robustness of our model to the choice of the NMS window hyperparameter ($\rho$), 
we perform further evaluation for the source-separtion experiment (See Table~\ref{tab:sep}) with (i) varying values for $\rho$, and (ii) using an oracle
that determines the optimal $\rho$ for every talking head from the ground truth bounding box size, instead of a fixed $\rho$.
The results are shown in Figure~\ref{fig:nms_oracle}. 

The experiment shows that very small or large constant NMS windows perform worse.
With qualitative inspection, we observe that too small values of $\rho$ lead to duplicate detections, while large
ones lead to merging instances.
However the oracle method, which has an adaptive window, does not obtain a significant improvement.

\daffy{in the rebuttal we also had the following regarding detection:
For the head detection experiments,  does not affect our method: Constant windows of size 100, 150
as well as the oracle, all result in 98.2\% and 99.7\% acc. on LRS2 and LRS3 respectively (expected, as there is only one speaker).
}

\section{Non-human speakers experiments} \label{app:non-human}
In this section, we provide more details about the dataset and evaluation on videos of non-human speakers.

\xpar{Unlabeled training sets.}
As training data for the non-human speakers experiments we used episodes of the \emph{The Simpsons} and
\emph{Sesame Street} shows found on YouTube.
The training sets we collected consist of approximately 48 hours for \emph{The Simpsons} (from seasons 11 to 31) 
and 53 hours of video for \emph{Sesame Street} (taken from playlists of the official YouTube channel
for several episode collections, as well as from playlists for characters Elmo, Cookie Monster,
Bert, Ernie, Abby, Grover, Rosita, Big Bird, Oscar, The Count, Kermit, and Zoe). The only processing
we perform on the original clips is splitting them into scenes by using the off-the-shelf package
\texttt{scenedetect}, so as to avoid clips with scene transitions.
We emphasize that no other preprocessing such as Voice Activity Detection or filtering out of title
frames was
performed; we trained our models in this raw, potentially noisy data.
We observed that clearly visible talking heads appear much more often in Simpsons episodes, compared to Sesame Street.
The latter also contains actual humans.
Moreover, the puppets used for the show are manually moved and there is only approximate correspondence in the
timing of movement with the corresponding speech, whereas
the head and mouth animations in Simpsons are temporally aligned with the speech. 
All of these factors make the training on examples from Simpsons significantly easier.


\xpar{Annotated test sets.}
To create the two test sets summarised in Table~\ref{tab:datasets}, we manually annotated
clips from held-out subsets, using the VIA annotator~\cite{dutta2019vgg}. 
There is no episode overlap between the training and test sets. 
We asked human labelers (three computer vision researchers) to annotate the active speaker in randomly chosen clips, including bounding boxes around the heads, in a small number of frames per clip. 
We note that in this case the character is not physically generating the sound; our goal is to reproduce these human judgements about which is the speaker (e.g., the ventriloquism effect for puppets).
For the \emph{multi-speaker} we also include negative samples that can be 
either non-speaking faces (those are the majority and we believe harder negatives) 
or frames not involving any characters, title/credit sequences, etc.
The ratio of positive and negative frames is approximately 1:1.

We include both {\em single-head} examples where only one speaker is in view (for a comparison to face detection methods on the localization task), and {\em multi-head} with multiple potential speakers for active speaker detection.

\xpar{Training details.} We trained separate models for the \emph{Simpsons} and the \emph{Sesame Street} experiments,
initialized from the best performing models trained on LRS2.

\xpar{Using off-the-shelf detectors and SyncNet.}
Face detectors are a key component of many speech understanding systems, such as active speaker
detection pipelines~\cite{Chung16a}, as well as for curating speech datasets~\cite{Chung16a,Afouras18d,Nagrani17,ephrat2018looking}. Here we investigate in more depth whether these off-the-shelf methods would also apply to non-human speakers in our dataset.
As described in Table~\ref{tab:cartoons}, we confirmed that an off-the-shelf face detector, RetinaFace~\cite{deng2019retinaface}, obtains poor average precision on these videos. 
In practice, correct face detections are poorly ranked and inconsistent frame to frame; thus it is
difficult to obtain them without introducing large numbers of false positives.
This behavior is expected, since these models have been trained on a different domain (human faces).
Here we provide qualitative examples of the detector's behavior (please see the video results), and a comparison to our self-supervised model's results.

Likewise, we also tried using SyncNet~\cite{Chung16a} as a baseline for the active speaker detection (ASD) task.
However, running this system out-of-the box failed.
This is because ASD with SyncNet is based on a multi-model pipeline:
first face detections are extracted with an SSD~\cite{Liu16} detector and heuristically stitched
into face tracks; SyncNet is then run to ASD on top of these face tracks. 
Since the face detector very rarely returns correct detections, producing virtually no face tracks, the model's later
steps were consistently incorrect.


\xpar{Extra non-human source-separation experiments.}
We also trained models to perform source-separation and speech enhancement on the \emph{Simpsons} data. 
For this we created synthetic videos with the mix-and-separate procedure. 
The separation model and training setting is the same as in the human speaker 
experiments as described in Sections~\ref{subsec:multispk} and \ref{subsec:trainingdetails}. 
We initialized the separation weights from the ones trained on LRS2.

We provide qualitative video results on our webpage.
In these, we demonstrate how our model uses the learned audio-visual objects to: i) successfully separate the voices of characters in multi-speaker clips; ii) handle challenging synthetic mixtures of the same character (e.g. Marge-Marge, Homer-Homer); iii) remove background noise and music.

\clearpage

\section{Architecture details.} \label{app:architecture_details}
\begin{figure}[h!]
\centering
\includegraphics[width=\textwidth]{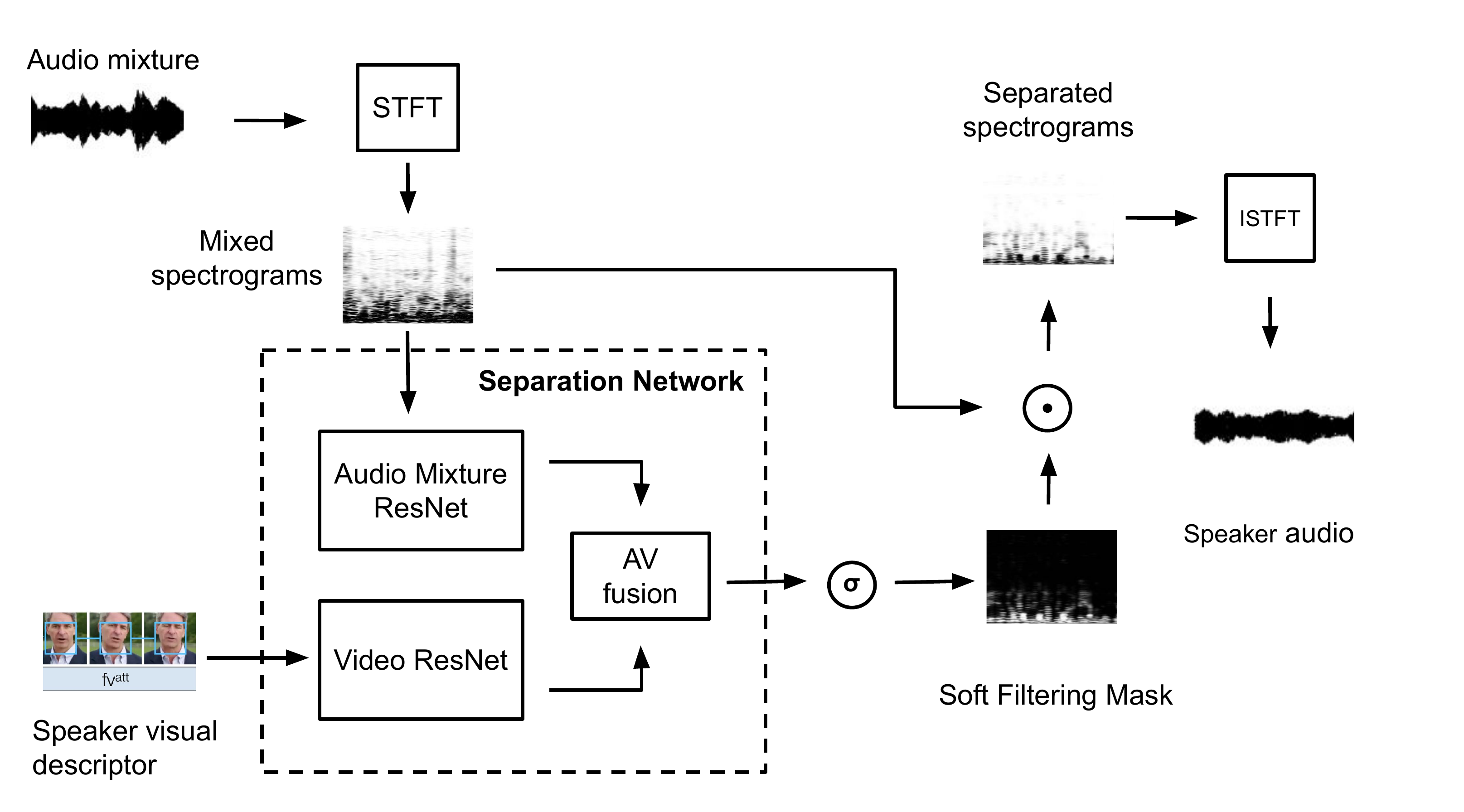}
\caption{ Separation network architecture. This is a detailed version of Figure~\ref{fig:sep}.
} 
\label{fig:sep_arch}

\end{figure}

In Table~\ref{tab:sync_arch}, we provide the full architecture for
the audio-visual synchronization module used for obtaining the attention maps.
In Figure~\ref{fig:sep_arch} and Table~\ref{tab:sep_arch} we give full architecture details for
the source separation module.

\begin{figure}[h!]
\centering
\includegraphics[width=\textwidth]{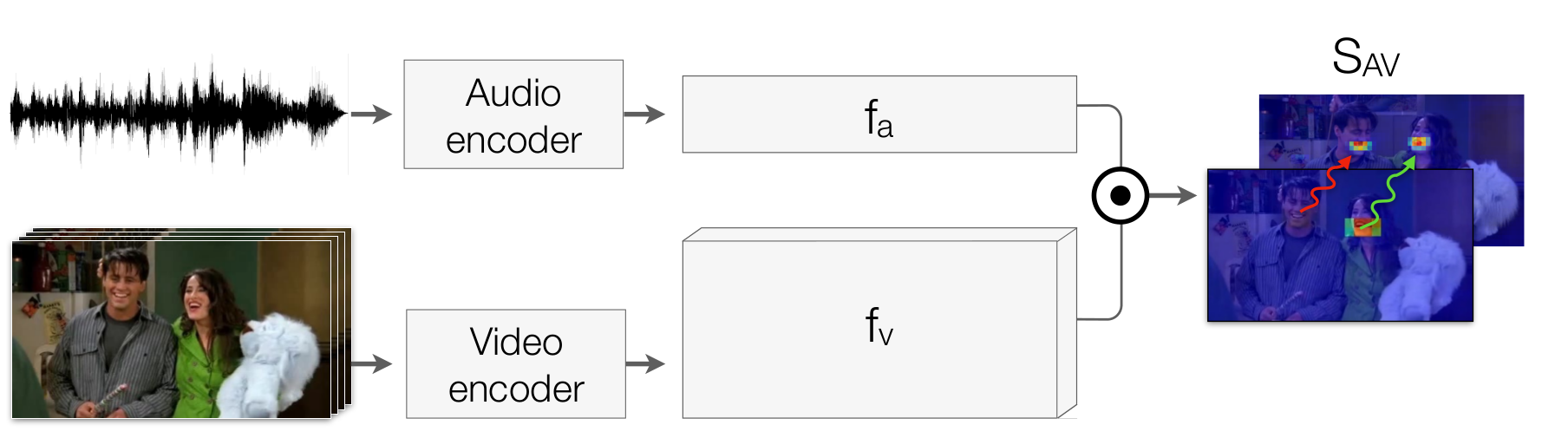}
\caption{ Synchronization network architecture. This is a part of Figure~\ref{fig:architecture}.
} 
\label{fig:sync_arch}
\end{figure}

\begin{table}[h!]
\setlength{\tabcolsep}{2pt}
\setlength\extrarowheight{0.5pt}
\caption{
  Architecture details for the audio-visual synchronization network, shown on Figure~\ref{fig:sync_arch}.
  We use a two-stream architecture similar to~\cite{Chung16a},
  containing a video and audio encoder that consume their respective modality and output
  embeddings in the same subspace.
  The embeddings are used to construct the audio-visual attention map $S_{av}$. 
  $K$ denotes kernel width and $S$ the strides (3 numbers for 3D convolutions and 2 for 2D convolutions).
  \textit{mp} denotes a max-pooling layer. 
  Batch Normalization and ReLU activation are added after every convolutional layer. 
  \textbf{Note:} To reduce clutter, $T$ was used in the paper instead of $T-4$ for the temporal
  dimension of the extracted embeddings.
}
\begin{minipage}[t]{0.40\hsize}\centering
\caption*{{\bf (a)} Audio Encoder}
\vspace{3pt}
\scriptsize
\begin{tabular}[t]{  l r c c r }
 \toprule
 Layer & \# filters & K & S   & Output  \\  
 \midrule
 input  & 1 &   - &  -             & $4T \times 80 $  \\  
 conv1  & 64 & (3,3)   & (1,2)     & $4T \times 40 $  \\  
 mp1   & -   & (3,1)   & (1,2)     & $4T \times 19 $  \\ 
 conv2 & 192 & (3,3)   & (1,1)     & $4T \times 19 $  \\ 
 mp2   & -   & (3,3)   & (2,2)     & $2T \times 9 $  \\ 
 conv3 & 256 & (3,3)   & (1,1)     & $2T \times 9 $  \\ 
 conv4 & 256 & (3,3)   & (1,1)     & $2T \times 9 $  \\
 conv5 & 256 & (3,3)   & (1,1)     & $2T \times 9 $  \\
 mp5   & -   & (3,3)   & (2,2)     & $T \times 4 $  \\
 conv6 & 512 & (4,4)   & (1,1)     & $T-4 \times 1 $  \\
 fc7   & 512  & (1,1)   & (1,1)     & $T-4 \times 1 $  \\
 fc8   & 1024 & (1,1)   & (1,1)     & $T-4 \times  1$  \\
 \bottomrule
\end{tabular}
\normalsize
\end{minipage}
\hfill
\begin{minipage}[t]{0.55\hsize}\centering
\caption*{{\bf (b)} Video Encoder}
\vspace{3pt}
\scriptsize
\begin{tabular}[t]{  l r r r r r }
 \toprule
 Layer & \# filters & K & S   & Output  \\  
 \midrule
 input  & 3 &   - &  -                   & $T\times H \times W$  \\  
 conv1    & 64 & (5,7,7) & (1,2,2)   & $T-4 \times \nicefrac{H}{2} \times \nicefrac{W}{2} $  \\  
 conv2 & 128 & (5,5)   & (2,2)     & $T-4 \times \nicefrac{H}{4} \times \nicefrac{W}{4} $  \\ 
 mp2   & - & (3,3)   & (2,2)       & $T-4   \times \nicefrac{H}{8} \times \nicefrac{W}{8} $  \\ 
 conv3 & 256 & (3,3)   & (1,1)     & $T-4 \times \nicefrac{H}{8} \times \nicefrac{W}{8} $  \\ 
 conv4 & 256 & (3,3)   & (1,1)     & $T-4 \times \nicefrac{H}{8} \times \nicefrac{W}{8} $  \\
 conv5 & 256 & (3,3)   & (1,1)     & $T-4 \times \nicefrac{H}{8} \times \nicefrac{W}{8} $  \\
 conv6 & 512 & (5,5)   & (1,1)     & $T-4 \times \nicefrac{H}{8} \times \nicefrac{W}{8} $  \\
 mp6   & - & (3,3)   & (2,2)       & $T-4   \times \nicefrac{H}{16} \times \nicefrac{W}{16} $  \\
 fc7  & 512  & (1,1)   & (1,1)     & $T-4 \times \nicefrac{H}{16} \times \nicefrac{W}{16} $  \\
 fc8  & 1024 & (1,1)   & (1,1)     & $T-4 \times \nicefrac{H}{16} \times \nicefrac{W}{16} $  \\
 \bottomrule
\end{tabular}
\normalsize
\end{minipage}

\normalsize
\label{tab:sync_arch}
\vspace{-27pt}

\end{table}

\clearpage

\begin{table*}[!h]
\vspace{-20pt}
\setlength{\tabcolsep}{5pt}

\setlength\extrarowheight{0.5pt}

\scriptsize
\caption{Architecture details for the Separation Network, shown on Figure~\ref{fig:sep_arch}.
  The modules are described in detail in~\cite{Afouras19b} and include:
  a) A 1D ResNet that processes the local descriptors extracted for each speaker-object.
  In particular the descriptors are pooled from the conv6 layer of the
  Video Encoder shown on Table~\ref{tab:sync_arch}.
  b) A 1D ResNet that processes the spectrogram of the audio mixture.
  c) A BLSTM and two fully-connected layers that perform the modality fusion.
  Notation: {\textit K:}~Kernel width; {\textit S:}~Stride -- fractional strides denote transposed convolutions;
  All convolutional layers are depth-wise separable.
  Batch Normalization, ReLU activation and a shortcut connection are added after every convolutional layer. 
  \textbf{Note:} We also use the phase refining network described in~\cite{Afouras18} for enhancing the
  phase of the audio signal, which we omit here for simplicity. For details please refer to the original paper.
}
\vspace{-5pt}
\begin{minipage}[t]{0.45\hsize}\centering
\caption*{{\bf (a)} Video ResNet}
\vspace{3pt}
\scriptsize
\begin{tabular}[]{  l r c c  r }
 \toprule
 Layer & \# filters & K & S   & Output  \\  
 \midrule
 input  & 512 &   - &  -  & $T\times 1$  \\  
 fc0    & 1536 & (1,1)   &  (1,1) & $T\times 1$  \\  
 conv1-2 & 1536 & (5,1)   &  (2,1) & $T\times 1$  \\  
 conv3 & 1536 & (5,1) & (\nicefrac{1}{2},1)  & $2T\times 1$  \\  
 conv4-6 & 1536 & (5,1) & (1,1)   & $2T\times 1$  \\  
 conv7 & 1536 & (5,1) & (\nicefrac{1}{2},1)   & $4T \times 1$  \\  
 conv8-9 & 1536 & (5,1) & (1,1)   & $4T \times 1$  \\  
 fc10  & 256 & (1,1) & (1,1)   & $4T \times 1$  \\  
 \bottomrule
\end{tabular}
\normalsize
\end{minipage}
\hfill
\begin{minipage}[t]{0.50\hsize}\centering
\caption*{{\bf (b)} Audio Mixture ResNet}
\vspace{3pt}
\scriptsize
\begin{tabular}[]{  l r c c r }
 \toprule
 Layer & \# filters & K & S  & Out  \\  
 \midrule
 input  & 80 &   - &  - & $T\times 1$  \\  
 fc0   & 1536 & (1,1) & (1,1)   & $4T \times 1$  \\  
 conv{1-5} & 1536 & (5,1) & (1,1)   & $4T \times 1$  \\  
 fc6   & 256 & (1,1) & (1,1)  & $4T \times 1$  \\  
 \bottomrule
\end{tabular}
\normalsize
\vspace{20pt}
\caption*{{\bf (c)} AV Fusion Network}
\vspace{3pt}
\scriptsize
\begin{tabular}{  l r r  }
 \toprule
 Layer & \# filters & Out  \\  
 \midrule
 input  & 512  & $4T\times 1$  \\  
 BLSTM   & 400 & $4T \times 1$  \\  
 fc1     & 600 & $4T \times 1$  \\  
 fc2     & 600 & $4T \times 1$  \\  
 fc\_mask     & F & $4T \times F$   \\  
 \bottomrule
\end{tabular}
\normalsize
\end{minipage}

\normalsize
\label{tab:sep_arch}
\vspace{-27pt}

\end{table*}

\end{subappendices}
\end{document}